\documentclass[letterpaper, 10 pt, journal, twoside]{ieeetran}
\usepackage{graphics} 
\usepackage{epsfig} 
\usepackage{times} 
\usepackage{amsmath} 
\usepackage{amssymb}  
\usepackage{multirow}
\usepackage{subfigure}
\usepackage{wrapfig}
\usepackage{booktabs} 
\usepackage{paralist}
\usepackage{color}
\usepackage{cite}
\usepackage{xspace}

\begin{document}
%
\title{ICM-3D: Instantiated Category Modeling for 3D Instance Segmentation}
%
%
%

\author{Ruihang Chu, Yukang Chen, Tao Kong, Lu Qi and Lei Li%
\thanks{Manuscript received: April, 18, 2021; Revised July, 20, 2021; Accepted August 19, 2021. This paper was recommended for publication by Editor Cesar Cadena upon evaluation of the Associate Editor and Reviewers' comments. (\textit{Corresponding author: Tao Kong.)}}
\thanks{Ruihang Chu, Yukang Chen, Lu Qi are with the Department of Computer Science and Engineering, The Chinese University of Hong Kong (e-mail: rhchu@cse.cuhk.edu.hk; yukangchen@cse.cuhk.edu.hk; luqi@cse.cuhk.edu.hk).}
\thanks{Tao Kong is with ByteDance AI Lab (e-mail: kongtao@bytedance.com).}%
\thanks{Lei Li is with the Computer Science Department, The University of California Santa Barbara (e-mail: lilei@cs.ucsb.edu). Lei Li is not supported by any funding for this work.}%
\thanks{Digital Object Identifier (DOI): see top of this page.}
}
\newcommand{\method}{{ICM-3D}\xspace}
%
%

\markboth{IEEE Robotics and Automation Letters. Preprint Version. Accepted August, 2021}
{Chu \MakeLowercase{\textit{et al.}}: ICM-3D: Instantiated Category Modeling for 3D Instance Segmentation} 

%



\maketitle

\begin{abstract}
Separating 3D point clouds into individual instances is an important task for 3D vision. 
It is challenging due to the unknown and varying number of instances in a scene. 
Existing deep learning based works focus on a two-step pipeline: first learn a feature embedding and then cluster the points. 
Such a two-step pipeline leads to disconnected intermediate objectives. 
In this paper, we propose an integrated reformulation of 3D instance segmentation as a per-point classification problem.
We propose \method, a single-step method to segment 3D instances via instantiated categorization. 
The augmented category information is automatically constructed from 3D spatial positions. 
We conduct extensive experiments to verify the effectiveness of \method and show that it obtains inspiring performance across multiple frameworks, backbones and benchmarks.
\end{abstract}


%
\IEEEpeerreviewmaketitle

\section{Introduction}
\begin{figure}[t]
\begin{center}
\subfigure[Embedding learning \quad \quad \quad \quad]{\includegraphics[width=0.52\linewidth]{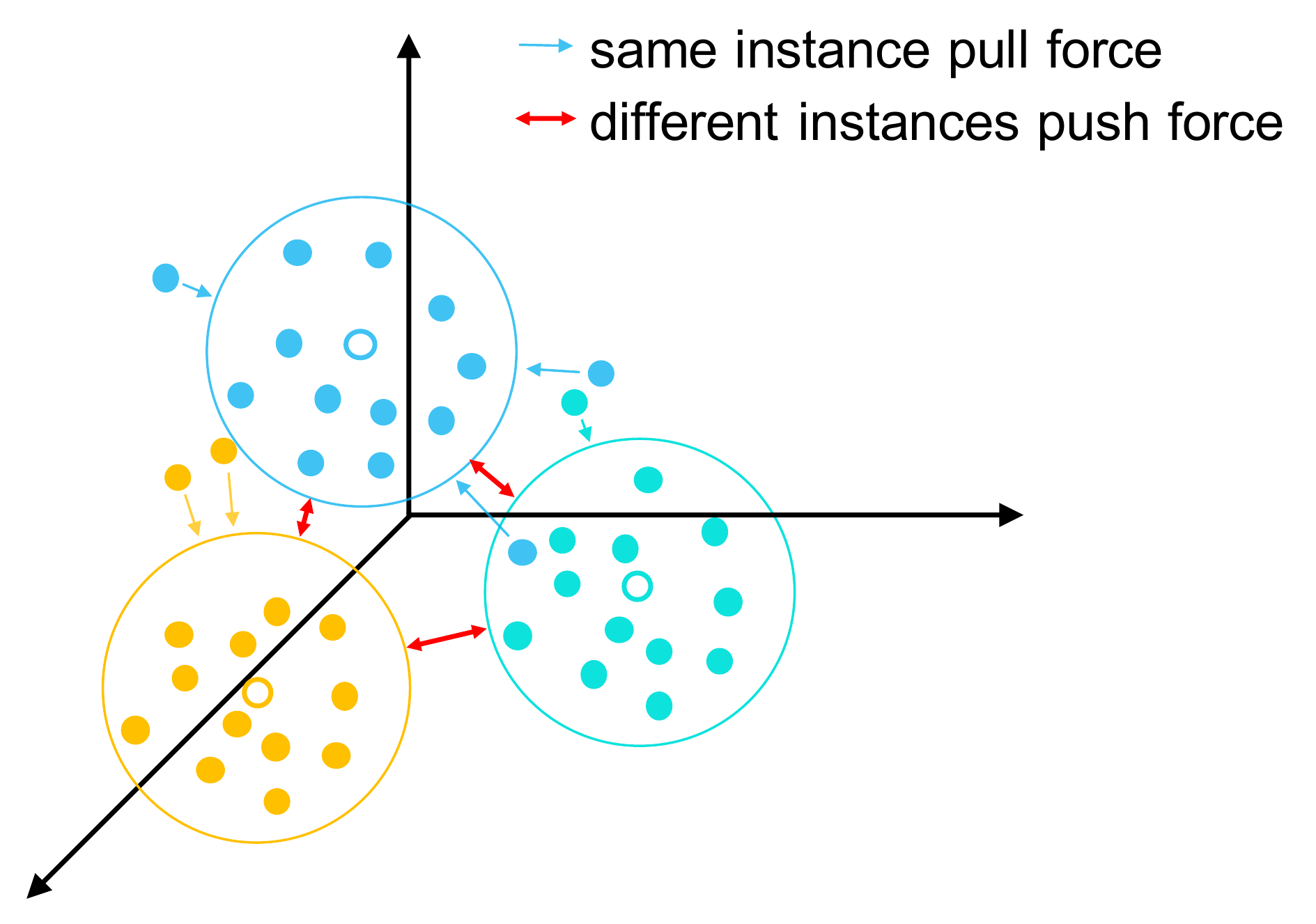}}
\subfigure[Instance category modeling]{\includegraphics[width=0.44\linewidth]{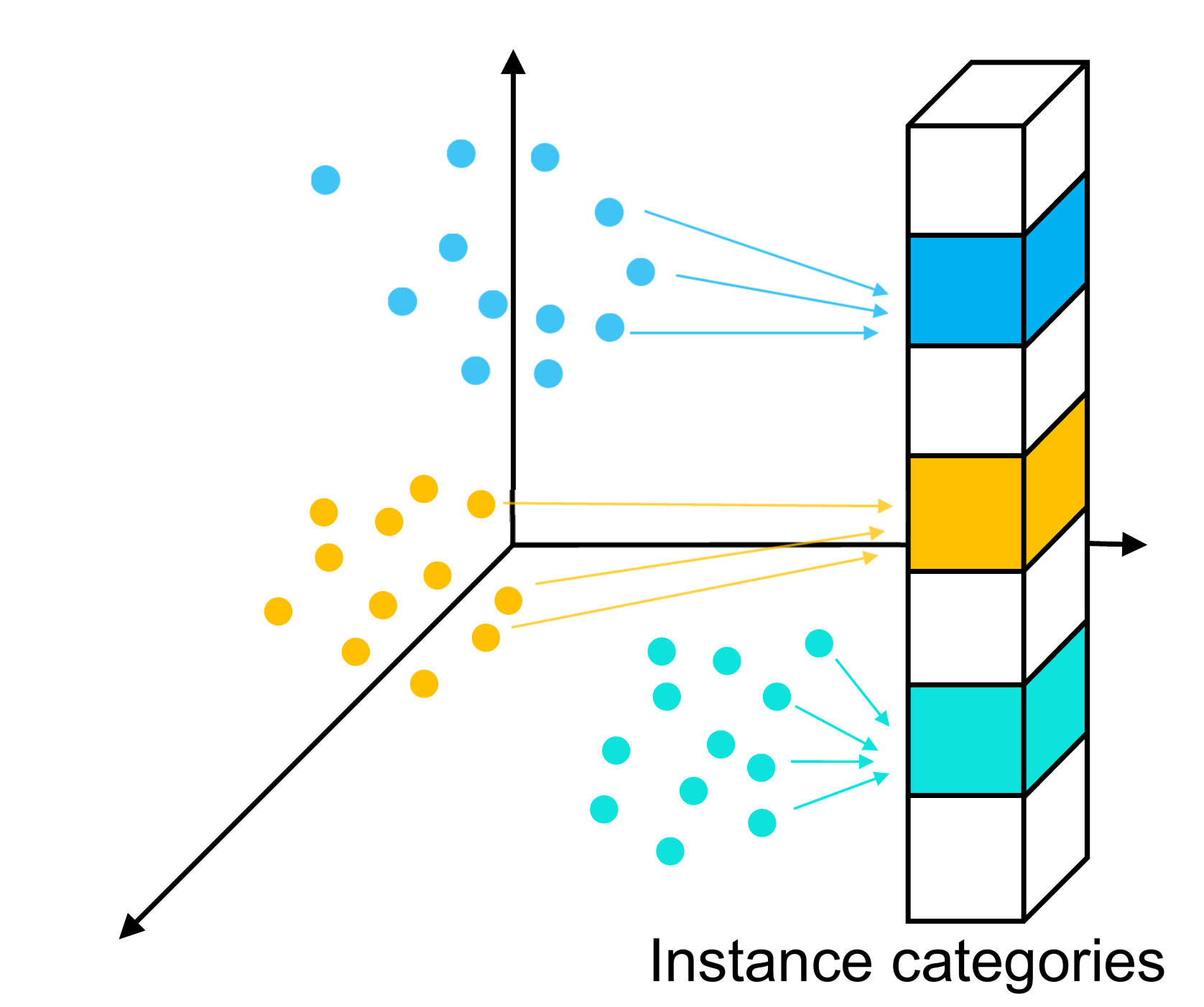}}
\caption{Comparison of the embedding learning method (a) and instantiated category modeling (\method) (b). 
(a) Embedding learning method uses discriminative loss to encourage points of the same instance to lie close together while different instances are separated, and adopts a clustering algorithm to obtain the final results;
(b) \method directly classifies each point to an instance category in the output space.
}
\label{fig:intro}
\vspace{-15pt}
\end{center}
\end{figure}

%
%
%
%
\label{sec:intro}
\IEEEPARstart{3}{D} point cloud instance segmentation~\cite{hou20193d,yi2019gspn,yang2019learningobject,wang2018sgpn,wang2019associatively,lahoud20193d,jiang2020pointgroup,han2020occuseg,zhang2021point,Engelmann_2020_CVPR,he2020instance} is crucial for many applications including autonomous driving, manipulating robots, and augmented reality. It requires not only predicting a semantic label for each point (\textit{e.g}., chair, table, ...), but also assigning points to each instance. 
3D point cloud instance segmentation is challenging.
The formulation of semantic label prediction has been widely and well explored in the literature~\cite{qi2017pointnet++,thomas2019kpconv,graham20183d}. 
However, how to formulate the process of separating different instances has not been well defined. 
Until now, the dominant methods~\cite{wang2019associatively,lahoud20193d,jiang2020pointgroup,han2020occuseg} usually consider 3D instance segmentation as a two-step per-point embedding learning and subsequent clustering problem. 
First, each point is embedded in a latent feature space, such that points belonging to the same instance should be close, and points of different instances should be partible (Fig.~\ref{fig:intro} (a)). 
Second, points are grouped following a clustering strategy, such as by mean-shift~\cite{comaniciu2002mean} algorithm~\cite{wang2019associatively,lahoud20193d}, breadth-first search based on distance threshold~\cite{jiang2020pointgroup}, or graph node merging~\cite{han2020occuseg}. 
An important assumption here is that points belonging to the same instance should have very similar features. Therefore, they mainly focus on discriminative feature learning and point clustering.

Despite the remarkable progress these \textit{embed-and-cluster} methods have achieved, we observe two potential issues in such a paradigm: 1) Instability: at each training iteration, the relative pair-wise feature distances are mainly optimized and each change usually influences others. As a result, the training objective becomes conceptually unstable and even ambiguous; 2) Inconsistency: at inference phase, clustering is a rule-making process that converts pair-wise distances to determined instance separations. Often it first requires an empirical distance threshold~\cite{wang2018sgpn}, or to estimate unknown instance centers as metrical reference~\cite{wang2019associatively,lahoud20193d}. This heuristic algorithm brings disconnected intermediate objectives between the training and inference.
Though some work tried to make the learned embedding feature more discriminative~\cite{lahoud20193d}, or proposed advanced clustering algorithms\cite{jiang2020pointgroup,han2020occuseg} specified for point clouds, the essential limitation is that \textit{there is no explicit and generalizable criterion to separate points belonging to different instances}. 

In this paper, we believe a more direct solution is to project different 3D instances to different spaces. If these instance-wise spaces are fixed, a classification objective could be naturally used for instance separation.
However, such a classification paradigm is usually specific to 3D semantic segmentation, since semantic classes are constant while the instance number is not easy to be predefined.

The key idea of our method is classify each point to a modeled \textit{instance category}.
Points belonging to the same instance will be projected to the identical instance category and otherwise be explicitly separated, as shown in Fig.~\ref{fig:intro} (b).
For training, a simple classification loss function is employed, converting instance segmentation into a classification-solvable problem. So here the key bottleneck is how to define the instance category, which must have fixed quantity and be unique for different instances.

We find the spatial position can naturally serve as the reference for defining the instance category, because instances scattered in 3D space rarely overlap with each other, and different instances occupy different spaces. Specifically, we uniformly quantify the 3D space by cubes based on spatial positions. 
Each cube serves as an instance category reference. During the training, we predict the instance category of each point by classifying this point to the cube reference where its instance centroid is located.
Moreover, we propose a novel \textit{project paradigm} that transforms direct instance category prediction to predicting it according to 3D projections.
We demonstrate this simplified projection gives equivalent modeling capacity, while it is much more computation and memory efficient.

This classification-based paradigm potentially has advantages to alleviate two aforementioned issues in the \textit{embed-and-cluster} pipeline: 1) Instability: points will be classified into fixed instance categories. Their training objectives are explicit and independent, out of the restriction of their mutual influence from relative distance enforcement, thus mitigating the training instability; 2) Inconsistency: we discard the heuristic clustering post-processing. The network workflow becomes single-step, where the training and inference objectives are both instance category prediction and keep consistent.
Our major contributions are as follows:

\begin{itemize}
    \item We propose an instantiated category modeling paradigm (\method) for 3D instance segmentation, which reformulates this complex task as a per-point classification problem. It not only gives an explicit and generalizable criterion for separating instances, but also eliminates additional clustering operations in previous work.
    \item  Our proposed \method can be seamlessly integrated into any neural network backbone to enable an end-to-end differentiable framework. We present two equivalent implementation forms to verify the instantiated category modeling paradigm. 
    \item We validate the effectiveness of \method across different frameworks, backbones and datasets, demonstrating its versatility. 
\end{itemize}

\section{Related Work}
\textbf{2D Image Instance Segmentation.}
2D instance segmentation was investigated ahead of 3D counterpart owing to the regular image structure and powerful 2D convolution neural networks. The major research line can be roughly grouped into two categories. The first family generate proposal boxes that contain objects and further classify pixels inside each box as objects or background (called top-down paradigm)~\cite{he2017maskrcnn,li2017fcis}. 
The second family are bottom-up methods~\cite{newell2017associative,gao2019ssap}, which learn an embedding vector for each pixel and apply a clustering at embedding space as post-processing. A similar work with our \method is SOLO~\cite{wang2019solo} as it refers to object locations and sizes for pixel classification. The  difference between two methods are detailed in Sec.~\ref{sec:solo}.

\textbf{Object Instance Segmentation.} A series of approaches\cite{pham2018scenecut,xie2021unseen,xiang2020learning,xie2021rice,DBLP:conf/icra/SuchiPFV19,potapova2014incremental} are targeted to segment 
objects for robot tasks.
Because discovering unseen objects is an important skill for robots, many of them aim to train on non-photorealistic RGB-D images while transfer well to real-world data. They succeed through handling RGB and depth separately~\cite{xie2021unseen}, effectively embedding synthetic RGB-D features~\cite{xiang2020learning}, and refining instance masks~\cite{xie2021rice}. We also show \method's generalizability on cluttered robotic scene dataset.

\textbf{3D Point Cloud Instance Segmentation.} Similar to 2D methods, solutions to 3D instance segmentation could be divided into the top-down and the bottom-up paradigm. The instance outputs of top-down approaches \cite{hou20193d,yi2019gspn,yang2019learningobject} are represented by inferred 3D instance boxes and the binary mask inside each box, where accurately predicting boxes is the key point.
To this end, 3D-SIS \cite{hou20193d} and GSPN \cite{yi2019gspn} generate boxes from redundant proposals, while 3D-BoNet \cite{yang2019learningobject} directly regresses and regularizes instance boxes.
Since the pruning of invalid boxes is computationally inefficient,
recent approaches prefer bottom-up paradigms \cite{wang2018sgpn,wang2019associatively,lahoud20193d,zhang2021point,han2020occuseg,jiang2020pointgroup,Engelmann_2020_CVPR}. These methods formulate 3D instance segmentation as a two-step \textit{embed-and-cluster} problem.
To boost network accuracy, ASIS \cite{wang2019associatively} leverages the complementary nature of semantic segmentation and instance segmentation, MTML \cite{lahoud20193d} proposes a multi-task metric learning strategy, enhanced by better clustering algorithms~\cite{jiang2020pointgroup} and loss functions\cite{han2020occuseg}. 
In our \method, the training objective is to classify each point to an instance category, and the classification results can be directly used as inferred instance output.
It requires neither bounding box detection nor clustering post-processing.

\section{Method}
In this section, we begin by briefly reviewing the widely-used \textit{embed-and-cluster} paradigm for 3D instance segmentation. Then we introduce the notation of \textit{instance category} and show how to formulate the 3D instance segmentation into a per-point classification problem. Finally, we describe how to seamlessly integrate it into the off-the-shelf networks.

\subsection{Review of Embed-and-cluster Pipeline}
The dominant \textit{embed-and-cluster} methods leverage that points belonging to the same object instance should have similar features while different instances have different features. More specifically, 
given a point cloud $\mathcal{P}$ of $N_p$ points, a deep point cloud learning network 
(\textit{e.g.}, PointNet++ \cite{qi2017pointnet++})
projects points to an embedding space, generating a feature matrix $\mathcal{F}\in\mathbb{R}^{N_p\times N_c}$, where $N_c$ denotes the embedding dimension. 
Taking the widely used discriminative loss function \cite{wang2018sgpn,wang2019associatively} for example, its training objective is formulated as:
\begin{align}
    \label{equ:similar}
    & \mathop{\min}\limits_{\mathcal{F}}\; \frac{1}{K}\sum_{k=1}^{K}\frac{1}{N_k}\sum_{\mathcal{F}_i\in I_k}[D(\mu^k, \mathcal{F}_i) - \delta^v]^2_+\\
    \label{equ:dissimilar}
    & \mathop{\min}\limits_{\mathcal{F}}\; \frac{1}{K(K-1)}\mathop{\sum_{k_A=1}^{K}\sum_{k_B=1}^{K}}\limits_{k_A\ne k_B}[2
    \delta^d-D(\mu^{k_A}, \mu^{k_B})]^2_+
\end{align}
where $K$ is the number of ground-truth instances; $N_k$ is the number of points in $k$-th instance $I_k$; $\mathcal{F}_i$ is the embedding of the $i^{th}$ point; $\mu^k$ is the mean embedding of the $k$-th instance;
$\delta^v$ and $\delta^d$ are predefined margins; $D(\cdot)$ means the distance and $[x]_+=max(0, x)$ means the hinge. 

As the model converges, the output features for the same instance should be similar~(Eq.~\eqref{equ:similar}), while features for different instances should be dissimilar~(Eq.~\eqref{equ:dissimilar}). 
During the inference, the final instance labels could be obtained by clustering the output features $\mathcal{F}$ using a conventional clustering algorithm, such as mean-shift~\cite{comaniciu2002mean}. 

From such a pipeline it is observed that the model is aware of the ground-truth instance mean embedding (\textit{i.e.}, $\mu^k$) during training. While at inference stage, $\mu^k$ is replaced by the estimated mean embedding of a cluster. Further, the estimation is effected by the clustering algorithm itself and distance margins (\textit{i.e.}, $\delta^v$ and $\delta^d$) in training stage.

\subsection{Instantiated Category Modeling}
Here we consider separating different instances from another perspective: If the model can predict the \textit{instance category} for each point, then different instances could be naturally separated, analogous to semantic category prediction in semantic segmentation.

\begin{figure}[t]
	\begin{center}
		\includegraphics[width=1\linewidth]{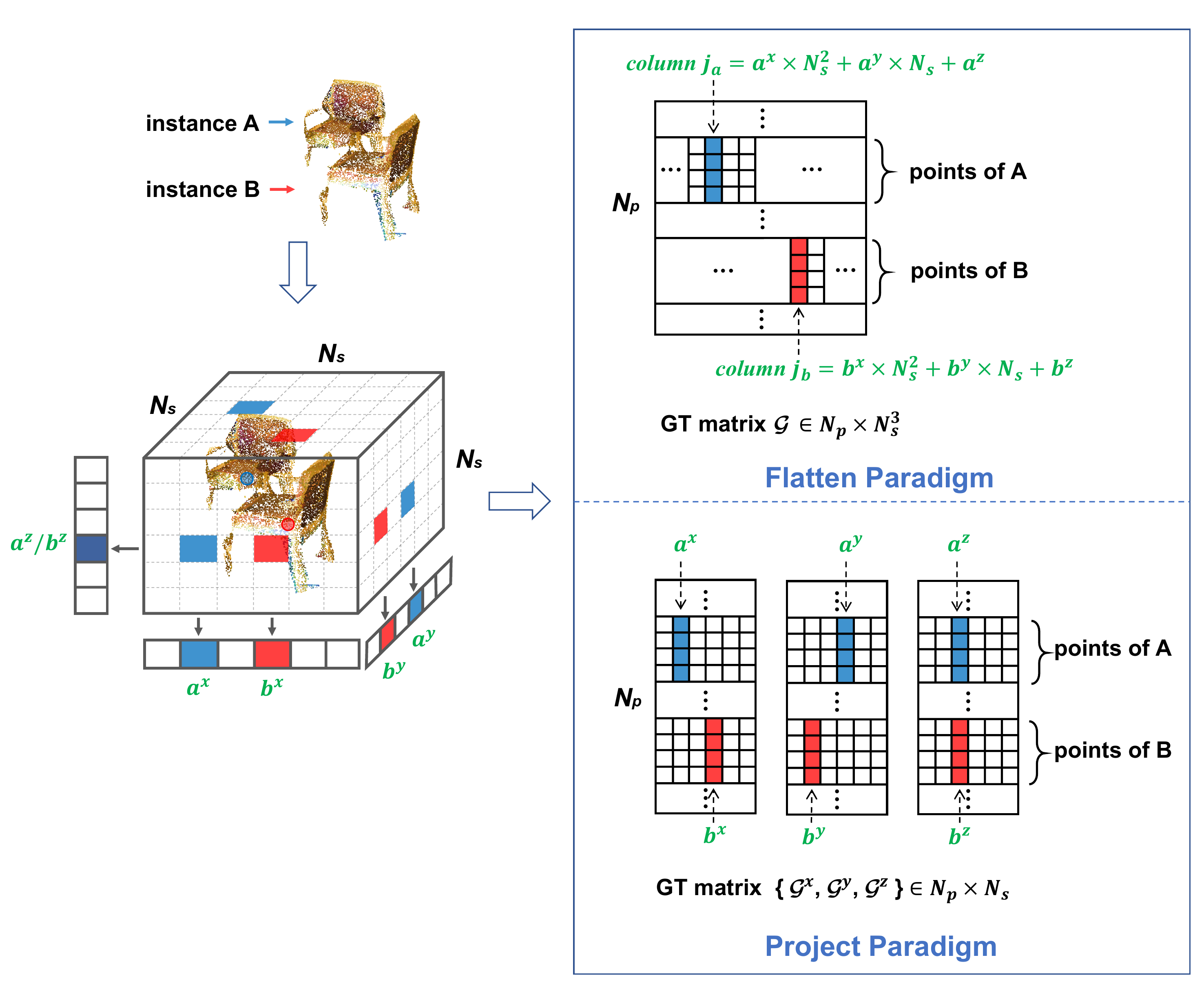}
	\end{center}
	\vspace{-10pt}
    \caption{Instantiated category modeling employs $N_s\times N_s\times N_s$ cubes to represent $N_s^3$ instance categories. Giving a point an instance label is to find its instance category, equaling to classifying the point to a cube. For instance A, we find the cube containing its centroid, denote the cube indexes along $X, Y, Z$ axis as $a^x$, $a^y$ and $a^z$ respectively. We classify points of instance $A$ to cube $(a^x, a^y, a^z)$ by:
    1) in the flatten paradigm, the column $j_a$ of their feature vectors is expected to be 1, denoted by blue in $\mathcal{G}$; 2) in the project paradigm, the column $a^x$, $a^y$, $a^z$ respectively in $\mathcal{G}^x$, $\mathcal{G}^y$, $\mathcal{G}^z$ should be 1.}
	\label{fig:method}
	\vspace{-10pt}
\end{figure}

\textit{How to define instance categories}? Fundamentally, it requires that each category represents a unique object instance, and meanwhile the category amount should be irrelevant to the varying instance number. Drawing upon the property of point cloud data, we find the spatial position can serve as the reference. Since instances scattered in 3D space rarely overlap with each other, different instances occupy different spaces.
Therefore, it is possible to establish a one-to-one correspondence between an instance category and a form of position representation. 
We uniformly divide the 3D space into $N_s\times N_s\times N_s$ small cubes and match each cube with one instance category. 
The rule is that a cube can only represent an instance whose centroid locates in the cube inner. Here an instance's centroid is denoted by the mean 3D coordinates of owned points.
Up to this point, the solution becomes very simple: for each point, we find the cube that contains the centroid of its respective instance, and then classify this point to this cube.
Two implementation choices are provided here, 
as \textit{flatten paradigm} and \textit{project paradigm}, respectively. 

\textbf{Flatten paradigm.}
In the flatten paradigm, we let the output of each point to be $N_c$ dimension, where $N_c=N_s^3$.
Naturally the $j^{th}$ output channel will correspond to cube ($x,y,z$) by letting $j$ equal the flattened cube index (\textit{i.e.}, $x\cdot N_s^2 + y\cdot N_s + z$).
For the $i^{th}$ point, suppose its corresponding instance's centroid is in cube ($i^x,i^y,i^z$), the target (ground-truth) matrix $\mathcal{G}\in\mathbb{R}^{N_p\times N_s^3}$ can be formulated as

\begin{equation}
    \label{equ:nodecouple}
    \mathcal{G}_{ij} = 1 \quad
    \mathrm{s.t.}\; j = i^x\cdot N_s^2 + i^y\cdot N_s + i^z.
\end{equation}

where $\mathcal{G}_{ij}$ is an element on the $i^{th}$ row, the $j^{th}$ column of $\mathcal{G}$. As shown in Fig.~\ref{fig:method}, we first find the cubes corresponding to instance $A$ and $B$, denoted as ($a^x, a^y, a^z$) and ($b^x, b^y, b^z$) respectively. 
For points of instance $A$, the column $j_a$ of their feature vectors is expected to be 1 (denoted with blue), other columns should be 0. Instance $B$ is represented in the same way.  
By projecting different instances to different positions,
points belonging to different object instances are directly separated according to their embedding values.

\textbf{Project paradigm.} As object instances are often located sparsely, expanding the channel number to $N_s^3$ is not worthwhile. 
We further provide a conceptually equal alternative, \textit{i.e.,} project paradigm. Specifically, we leverage the principle that a spatial position can be uniquely defined by its orthogonal projections along $X$, $Y$, $Z$ axis. 
Thus we generate three independent embeddings, each has $N_s$ channels that correspond to a cube index at one axis. 
Accordingly, the output space declines from $\mathbb{R}^{N_p\times N_s^3}$ to $\mathbb{R}^{N_p\times 3N_s}$. In this paradigm, assume that cube ($i^x,i^y,i^z$) represents the instance category of $i^{th}$ point, the target (ground-truth) matrix $\mathcal{G}$ can be written as
$
    \mathcal{G} = \{\mathcal{G}^{x},\mathcal{G}^{y},\mathcal{G}^{z}\}
$.
Taking $\mathcal{G}^{x}$ as an example, 
\begin{equation}
    \label{equ:decouplex}
    \mathcal{G}_{ij}^{x} = 1 \quad
    \mathrm{s.t.}\; j = i^x.
\end{equation}
Likewise, we let the rest of the elements in $\mathcal{G}$ be 0. The bottom branch of Fig.~\ref{fig:method} exhibits the target matrix $\mathcal{G}$ according to the cube indexes of instances.  

After defining the ground-truth feature matrix, the model can be trained under its supervision. For both implementation paradigms, our training objective is 
\begin{equation}
    \label{equ:objective}
    \mathop{\min}\limits_{\mathcal{F}}\; D(\mathcal{F}, \mathcal{G})
\end{equation}
\vspace{-10pt}

As each value in $\mathcal{G}$ is binary, Eq.~\eqref{equ:objective} suggests that the training objective can be simply trained by, but not limited to, commonly used classification loss functions. Compared with Eq.~\eqref{equ:similar} and Eq.~\eqref{equ:dissimilar}, it introduces no hyperparameters such as margins. 
During the inference, different instances can be naturally separated by their instance categories.

\subsection{Integrating \method into Networks}
\label{sec:networks}

\textbf{Network Architecture.} Given the point cloud $\mathcal{P}$ of $N_p$ points, we use a off-the-shelf backbone network for point cloud (\textit{e.g.,} PointNet++ \cite{qi2017pointnet++}) to output per-point local features, denoted as $\mathcal{F}_l\in\mathbb{R}^{N_p\times N_l}$, where $N_l$ is the channel number.
Then $\mathcal{F}_l\in\mathbb{R}^{N_p\times N_l}$ is directly mapped to $\mathcal{F}\in\mathbb{R}^{N_p\times N_c}$ via simple multi-layer perceptions (MLPs). 
For the flatten paradigm, we employ two-layer 32-channel MLPs
and end with a \texttt{sigmoid} operation, producing $\mathcal{F}\in\mathbb{R}^{N_p\times N_s^3}$. As for the project paradigm, the network proceeds with three independent branches to transform $\mathcal{F}_l$ to three feature matrices, denoted as $\mathcal{F}^x$, $\mathcal{F}^y$, $\mathcal{F}^z$ respectively, all with the shape $\mathbb{R}^{N_p\times N_s}$. 
Each branch is composed of three-layer 32-channel MLPs 
and a \texttt{sigmoid} layer. Here we use one more layer MLP than the flatten version, so as to more thoroughly decouple position information along $X$, $Y$, $Z$ axis.

Although we define $N_s^3$ instance categories by spatial cubes, only a few cubes contain object instance's centroid. It requires the network to select valid instance categories. We solve this problem by giving each cube a score. The higher the score, the more likely that the cube denotes an actual instance.
Specifically, the network in parallel appends an instance category scoring head. It transforms $\mathcal{F}_l\in\mathbb{R}^{N_p\times N_l}$ to a $N_s^3$-dimension vector. Each value 
in the vector represents a cube's score. 
As cube's validity is closely relevant to its local pattern, we propose to dynamically aggregate contextual features specified for sparse point cloud, making the learned local pattern adaptive to point densities and object sizes. We imitate the `set abstraction level' in PointNet++ \cite{qi2017pointnet++}, find the top 32 closest points around the center of each cube, and group their features via an average pooling layer. The output score is finally scaled to (0, 1) with a \texttt{sigmoid} operation. 

\textbf{Training.}
The training target for instance category prediction is illustrated in Eq.~\eqref{equ:objective}. For the flatten paradigm, the 
column number 
of output feature matrix 
is large, a large part of 
columns
are invalid instance categories and are meaningless. 
We define our instance category loss as 

\begin{equation}
\label{equ:l_cate}
    \mathcal{L}_{pcate} = \frac{1}{N_{pos}}\sum_{j}\textbf{P}_j^*D(\mathcal{F}_{ij}, \mathcal{G}_{ij})
\end{equation}
\vspace{-10pt}

where $N_{pos}$ denotes the number of valid cubes. $\textbf{P}_j^*$ is the indicator that corresponds to the $j^{th}$ 
column, being 1 if the $j^{th}$ cube is positive and 0 otherwise. 
For the project paradigm, we define 
$
    \mathcal{L}_{pcate} = D(\mathcal{F}^x_{ij}, \mathcal{G}_{ij}^{x})+D(\mathcal{F}^y_{ij}, \mathcal{G}_{ij}^{y})+D(\mathcal{F}^z_{ij}, \mathcal{G}_{ij}^{z})
$. For $D(\cdot,\cdot)$, we choose the stable Dice Loss~\cite{milletari2016vnet}
\begin{equation}
\label{equ:l_dice}
    \mathcal{L}_{Dice} = 1 - \frac{2\sum_{i,j}(\mathcal{F}_{ij}\cdot \mathcal{G}_{ij})}{\sum_{i,j}\mathcal{F}_{ij}^2 + \sum_{i,j}\mathcal{G}_{ij}^2}.
\end{equation}

The instance category scoring head is supervised by ground-truth cube scores. Following widely-used center sampling strategy as in \cite{kong2019foveabox}, we replace the centroid of an instance with a center region that is proportional to the object bounding box.
Accordingly, multiple cubes within an instance center region may be regarded as positive and each point can thus have more than one instance categories. Note that the redundant ones can be removed during the inference, as detailed later.
The benefits derived from this policy could be two-fold. First, the center region reflects the size and shape information of each instance, whereby the positive cube number will be adjusted automatically. Second, predicting multiple positive categories for one point effectively enhances the robustness and fault tolerance.
During the training, we employ a Binary Cross Entropy loss to classify each cube to either positive or negative, denoted as $\mathcal{L}_{score}$.

To generate semantic labels, we follow recent works \cite{yang2019learningobject,lahoud20193d}, pass backbone features with a MLP to learn per-point semantic labels, and train it with standard softmax cross-entropy loss. Denoting the semantic loss as $\mathcal{L}_{sem}$, our loss function $\mathcal{L}$ is computed as: $\mathcal{L} = \mathcal{L}_{score} + \mathcal{L}_{pcate} + \mathcal{L}_{sem}$. 

\textbf{Inference.} 
In the flatten paradigm, our network outputs the feature matrix ($\mathcal{F}\in\mathbb{R}^{N_p\times N_s^3}$), where each column $\mathcal{M}\in\mathbb{R}^{N_p}$ indicates the instance mask referred by the corresponding cube. 
For each column $\mathcal{M}$, 
the matrix elements are firstly binarized with a threshold of 0.5. Columns whose cube scores less than $t$ are regarded as negative ones and will be discarded. 
Finally, as the center sampling strategy brings redundant predictions, we utilize 
non-maximum suppression (NMS)~\cite{jiang2020pointgroup} to suppress predictions (columns) according to their scores and the mask IoUs. 
For the project paradigm, we obtain three feature matrices $\mathcal{F}^x$, $\mathcal{F}^y$, $\mathcal{F}^z\in \mathbb{R}^{N_p\times N_s}$. To obtain the corresponding instance mask of cube $j$, we find the $i^x$, $i^y$, $i^z$ column of $\mathcal{F}^x$, $\mathcal{F}^y$, $\mathcal{F}^z$ respectively and multiply them in an element-wise way,
where $j = i^x\cdot N_s^2 + i^y\cdot N_s + i^z$. The semantic class of an instance is the most occurring predicted label among all points it contains.

\section{EXPERIMENTS}

\subsection{Experimental Settings}
For the following experiments, we use the flatten paradigm, set $N_s=20$ and employ PointNet++ \cite{qi2017pointnet++} as the backbone network if not specified. 
The method is evaluated on two 
large-scale scene-level
dataset S3DIS~\cite{armeni20163d}, ScanNet V2~\cite{dai2017scannet}, a synthetic object-level dataset PartNet \cite{mo2019partnet}, and a cluttered object dataset OCID~\cite{DBLP:conf/icra/SuchiPFV19}.

\begin{table*}[!htbp]
\vspace{10pt}
\begin{minipage}{0.7\textwidth}
\caption{Instance segmentation results on S3DIS.
\method consistently achieves impressive results on multiple backbone networks.}
\vspace{-5pt}
\begin{center}
    \renewcommand\arraystretch{1.1}
		\setlength{\tabcolsep}{0.7mm}{
  \begin{tabular}{l|l|cccc|cccc}
    \toprule
    \multirow{2}{*}{Backbone} & \multirow{2}{*}{Method} & \multicolumn{4}{c|}{Test on Area5} & \multicolumn{4}{c}{Test on 6-fold c. v.} \\ 
    \cline{3-10}
    & & mCov & mWCov & mPrec & mRec & mCov & mWCov & mPrec & mRec \\
    \hline
    \multirow{2}{*}{PointNet} & SGPN~\cite{wang2018sgpn} & 32.7 & 35.5 & 36.0 & 28.7 & 37.9 & 47.8 & 38.2 & 31.2 \\
    & \method & \textbf{43.4}  & \textbf{44.6} & \textbf{45.1} & \textbf{39.5} & \textbf{46.9} & \textbf{50.1} & \textbf{54.2} & \textbf{43.0}\\
    \hline
    \multirow{3}{*}{PointNet++}
    & ASIS~\cite{wang2019associatively} & 44.6 & 47.8 & 55.3 & 42.4 & 51.2 & 55.1 & 63.6 & 47.5 \\
    & 3D-BoNet~\cite{yang2019learningobject} & - & - & - & - & - & - & 65.6 &47.6 \\
    & \method & \textbf{46.9} & \textbf{49.6} & \textbf{57.0} & \textbf{44.6} & \textbf{53.4} &  \textbf{57.1} & \textbf{65.9} & \textbf{49.6}\\
    \hline
    \multirow{2}{*}{PointConv} & ASIS & 44.7 &  48.1 & 55.3 & 43.2 & 51.5 & 55.1 & 63.8 & 48.0\\
    & \method & \textbf{47.0} & \textbf{49.8} & \textbf{57.4} & \textbf{45.0} & \textbf{53.9} & \textbf{57.3} & \textbf{65.9} & \textbf{49.8}\\
    \bottomrule
  \end{tabular}
  }
  \end{center}
  \label{table-s3dis}
\end{minipage}
\begin{minipage}{0.28\textwidth}
\caption{Instance segmentation results on ScanNet V2 validation set.}
	\begin{center}
		\renewcommand\arraystretch{1.11}
		\setlength{\tabcolsep}{0.7mm}{
\begin{tabular}{l|ccc}
\toprule
     Method & mAP & @50\% & @25\% \\
\hline
    SGPN~\cite{wang2018sgpn}& - & 11.3 & 22.2 \\
    3D-SIS~\cite{hou20193d} & - & 18.7 & 35.7 \\
    IAM~\cite{he2020instance} & - & - & 50.4 \\
    GSPN~\cite{yi2019gspn} & 19.3 & 37.8 & 53.4 \\
    3D-BoNet~\cite{yang2019learningobject} & 19.2 & 39.1 & 55.2 \\
    MTML~\cite{lahoud20193d}& 20.3& 40.2 & 55.4 \\
    PointGroup~\cite{jiang2020pointgroup}\ddag & 30.5 & 47.2 & 61.9 \\
    ICM-3D & \textbf{32.7} & \textbf{52.3} & \textbf{67.8} \\
\bottomrule
\end{tabular}}
 \end{center}
 \label{tab:scannet}
\end{minipage}
\end{table*}

\begin{table*}[t]
\vspace{-5pt}
\caption{Instance segmentation results on fine-grained-level benchmark of PartNet. \method achieves the best average score (mAP).}
\vspace{-5pt}
\begin{center}
    \renewcommand\arraystretch{1.1}
		\setlength{\tabcolsep}{1.1mm}{
  \begin{tabular}{l|c|ccccccccccccccccc}
    \toprule
     Method & mAP & Bed & Bott & Chair & Clock & Dish & Disp & Door & Ear & Fauc & Knife & Lamp & Micro & Frid & Stora & Table & Trash & Vase \\
    \hline
     SGPN~\cite{wang2018sgpn} & 29.5 & 11.8 & 45.1 & 19.4 & 18.2 & 38.3 & 78.8 & 15.4 & 35.9 & 37.8 & 38.3 & 14.4 & 32.7 & 18.2 & 21.5 & 14.6 & 24.9 & 36.5 \\
     GSPN~\cite{yi2019gspn} & - & - & - & - & - & - & - & - & - & - & - & 18.3 & - & - & - & 21.5 & 24.9 & - \\
     PartNet~\cite{mo2019partnet} & 36.6 & 15.0 & 48.6 & 29.0 & \textbf{32.3} & \textbf{53.3} & 80.1 & 17.2 & 39.4 & 44.7 & 45.8 & 18.7 & 34.8 & 26.5 & 27.5 & 23.9 & 33.7 & 52.0 \\
    PE~\cite{zhang2021point} & 39.8 & \textbf{26.2} & 50.7 & 34.7 & 30.2 & 50.0 & 82.0 & 25.7 & 43.2 & 55.6 & 44.4 & 20.3 & \textbf{37.0} & 31.1 & 34.2 & 25.5 & \textbf{37.7} & 47.6 \\
     \method & \textbf{46.6} & 20.7 & \textbf{55.3} & \textbf{46.9} & 29.9 & 52.8 & \textbf{86.7} & \textbf{28.9} & \textbf{45.7} & \textbf{56.3} & \textbf{52.1} & \textbf{51.0} & 32.3 & \textbf{49.3} & \textbf{55.5} & \textbf{40.0} & 29.5 & \textbf{59.8} \\
     \bottomrule
  \end{tabular}}
  \end{center}
\label{table-partnet}
\vspace{-10pt}
\end{table*}

\begin{table}[htbp]
\vspace{5pt}
\caption{Evaluation on ARID20 and YCB10 subsets of OCID.}
\begin{center}
    \renewcommand\arraystretch{1.0}
		\setlength{\tabcolsep}{2.3mm}{
  \begin{tabular}{l|ccc|ccc}
    \toprule
    \multirow{2}{*}{Method} &  \multicolumn{3}{c|}{Overlap} & \multicolumn{3}{c}{Boundary} \\ 
    \cline{2-7}
    & P & R & F & P & R & F \\
    \hline
    SCUT~\cite{pham2018scenecut} & 45.7 & 72.5 & 43.7 & 43.1 & 65.1 & 42.6\\
    V4R~\cite{potapova2014incremental} & 65.3 & 81.4 & 69.5 & 62.5 & 81.4 & 66.6\\
    UOIS-Net-3D~\cite{xie2021unseen} & \textbf{88.2} & \textbf{88.0} & 87.9 & 81.1 & \textbf{74.3} & 77.3\\
    \method & \textbf{88.2} & 87.9 & \textbf{88.2} & \textbf{81.3} & \textbf{74.3} & \textbf{77.7}\\
    \bottomrule
  \end{tabular}
  }
  \end{center}
  \label{tab:ocid}
  \vspace{-15pt}
\end{table}

\textbf{S3DIS} covers over 215 million points from 272 scanned 3D real scenes (\textit{e.g.,} office) in 6 large areas. For a fair comparison, we exactly adopt the training settings in~\cite{wang2018sgpn,wang2019associatively,yang2019learningobject} on input form (\textit{i.e.,} a 9-dim vector composed of RGB, XYZ and normalized coordinates of the room) and block-wise training.
The performance is evaluated on Area5 set and 6-fold cross validation, using the widely-used evaluation protocols as \cite{wang2018sgpn,wang2019associatively}, \textit{i.e.,} mCov, mWcov, mPrec and mRec.

\textbf{ScanNet V2} contains real 3D scenes reconstructed from rich-annotated RGB-D scans. We train our model on 1,201 scenes and implement evaluation on the validation set that consists of 312 scenes. We completely adopt the training and testing settings of \cite{yang2019learningobject}. The mean average precision (mAP) at an IoU threshold 0.25, 0.5 and averaged over the range [0.5:0.95:05] are used as the evaluation metric.

\textbf{PartNet} contains 573,585 part instances over 26,671 3D objects. The instance segmentation task is to segment every individual part from shape contexts.  
We follow all common training settings of original paper~\cite{mo2019partnet} and test our model on the most difficult 'fine-grained-level' benchmark by calculating mean Average Precision (mAP) on 17 object categories. 

\textbf{OCID} provides 2,346 challenging RGB-D images of cluttered scenes. We rate the quality of predicted object masks extensively by precision/recall/F-measure (P/R/F) metrics on Overlap and Boundary respectively as defined in~\cite{DBLP:conf/icra/SuchiPFV19,xie2021unseen}.

We train our model on a single Tesla-V100-16GB GPU, applying the Adam optimizer with default parameter settings. The learning rate is initialized as 0.001 and decreased by a factor of 0.5 after every 300k iterations. For all backbone networks and datasets, the model is trained for 100 epochs with batch size as 4. During training, the scaling factor of the center sampling strategy is set as 0.2 to supervise the instance category scoring head. The threshold to filter valid cubes at inference stage is set as 0.3 and NMS IoU is 0.3.

\subsection{Main Results}

\label{sec:s3dis}

\textbf{S3DIS.} We list other popular methods that adopt the same training settings (\textit{e.g.,} backbone, block-wise training) with \method for a fair comparison. 
Among these approaches, SGPN~\cite{wang2018sgpn} and ASIS~\cite{wang2019associatively} are embedding learning based methods and 3D-BoNet~\cite{yang2019learningobject} is a detection-based method. As shown in Tab.~\ref{table-s3dis}, \method outperforms them on all evaluation metrics.
SPGN indicates instances 
by the similarity score of each point pair, 
while our explicit point-level classification surpasses it by large margins. 
ASIS 
leverages semantic awareness to repel instances. \method gains +1.8\%$\sim$2.1\% mRec over ASIS with the same backbone. 
3D-BoNet predicts bounding boxes that benefit instance objectiveness. 
Compared with 3D-BoNet, \method still yields a +2\% mRec improvement.
Note that \method contains the fewest parameters among these works, \textit{i.e.,} a backbone with only a few MLPs.

To demonstrate the generality of \method, we conduct experiments on several widely-used point feature learning backbones, including MLP-based PointNet \cite{qi2017pointnet}, PointNet++ \cite{qi2017pointnet++}, and convolution-based PointConv \cite{wu2019pointconv}. 
The results show that \method achieves 
sustained advantage over other methods under various backbones (1\%$\sim$11.8\% mRec gains).

\begin{figure*}[t]
\vspace{10pt}
\begin{minipage}{0.58\textwidth}
	\begin{center}
	\vspace{5pt}
		\includegraphics[width=0.93\textwidth]{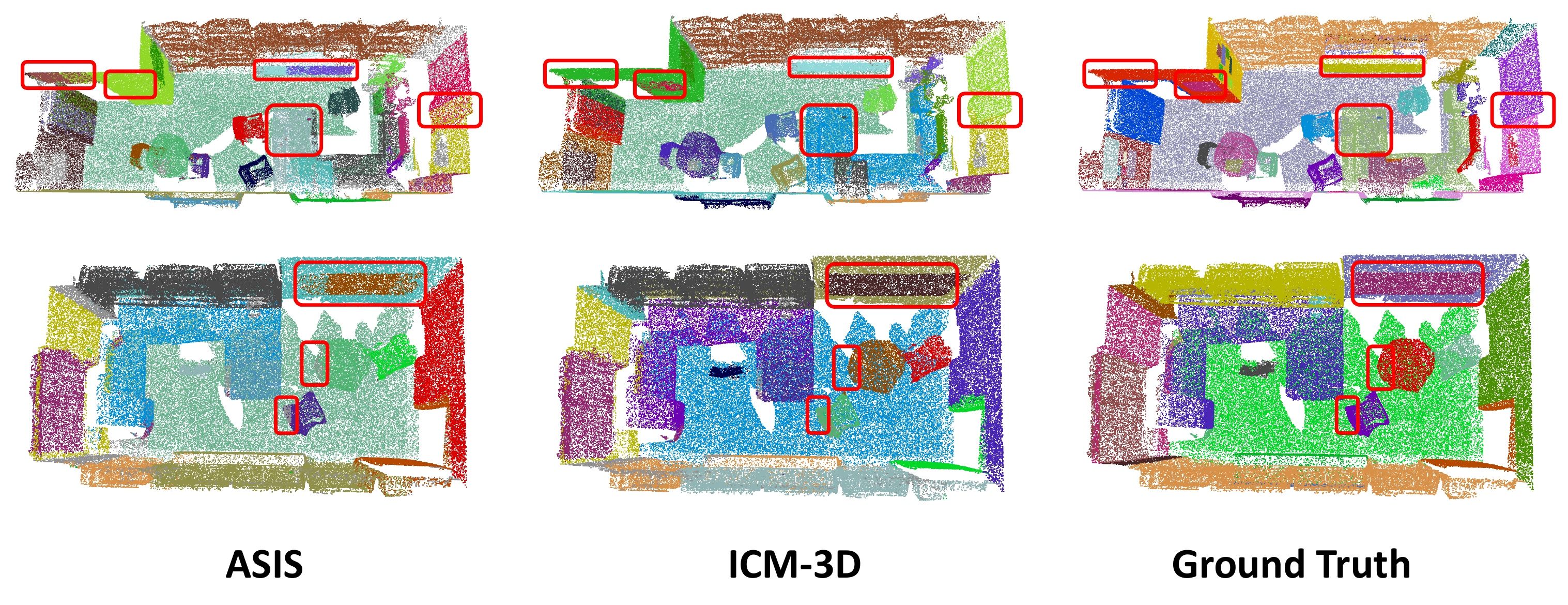}
	\end{center}
	\vspace{-10pt}
	\caption{Visualization results on S3DIS. 
	\method predicts more precise instance labels. The main differences are circled by red boxes.}
	\label{fig:s3dis_vis}
\end{minipage}
\hspace{5mm}
\begin{minipage}{0.37\textwidth}
	\begin{center}
		\includegraphics[width=0.98\textwidth]{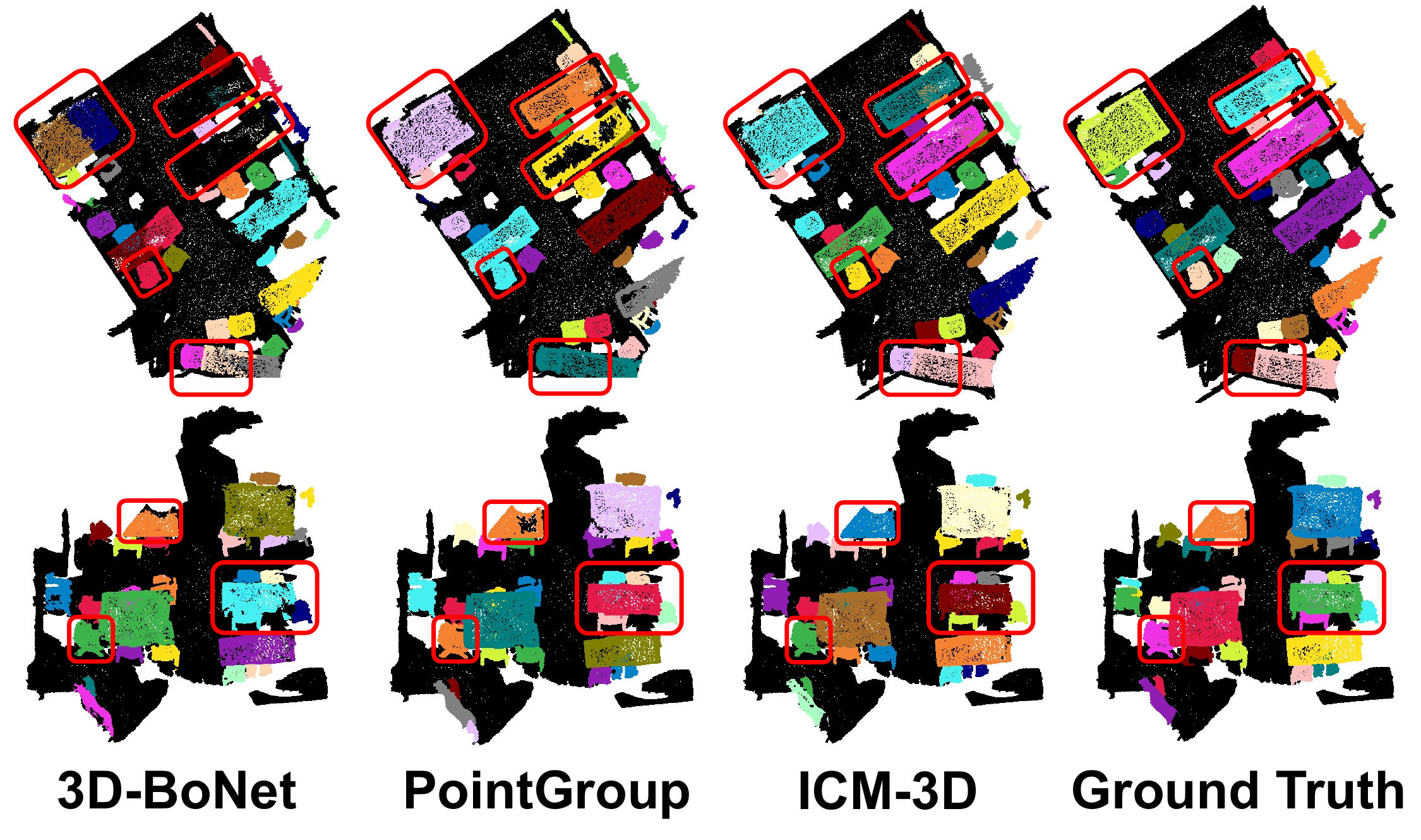}
	\end{center}
		\vspace{-10pt}
	\caption{Visualization results on ScanNet V2 validation set. Zoom in for better visualization.}
	\label{fig:scannet_vis}
\end{minipage}
\end{figure*}

We further introduce visualization comparisons in Fig.~\ref{fig:s3dis_vis}. 
It can be observed that \method produces more clean and robust instance predictions than ASIS, especially on object boundaries. This is because clustering-based approaches do not explicitly detect object boundaries, while \method directly classifies points to separated category sub-spaces and thereby ensures high objectiveness.

\textbf{ScanNet V2.} \method achieves inspiring performance on all evaluation metrics compared to other work in Tab.~\ref{tab:scannet}. Here we follow the experimental settings of 3D-BoNet~\cite{yang2019learningobject}, \textit{i.e.,} PointNet++ backbone with a separate SCN~\cite{graham20183d} network for per-point semantic prediction. 
For a fair comparison, we migrate the powerful PointGroup~\cite{jiang2020pointgroup} to the same PointNet++ based settings on its official repository, denoted by $\ddag$, which shows inferior accuracy than \method (-2.2\% mAP). The original PointGroup employs a much stronger backbone network along with significantly different training settings, compared to which \method achieves a bit lower mAP (-2.1\%). Examples of visualization comparison are exhibited in Fig.~\ref{fig:scannet_vis} to show the high quality of our instance predictions.

\begin{figure}[t]
	\vspace{-5pt}
	\begin{center}
		\includegraphics[width=1\linewidth]{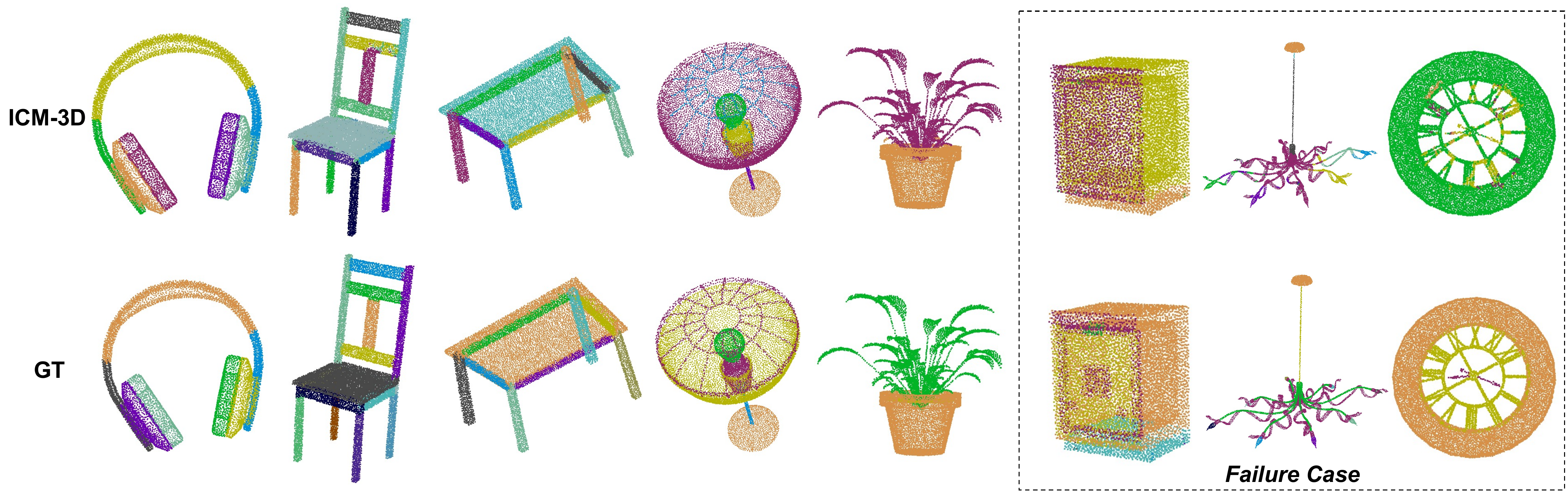}
	\end{center}
	\vspace{-10pt}
	\caption{Visualization results of ICM-3D on PartNet.}
	\label{fig:partnet_vis}
	\vspace{-10pt}
\end{figure}

\textbf{PartNet.} We compare \method with PE~\cite{zhang2021point}, PartNet \cite{mo2019partnet}, GSPN \cite{yi2019gspn} and SPGN \cite{wang2018sgpn} in Tab.~\ref{table-partnet}. All models adopt PointNet++~\cite{qi2017pointnet++} as the backbone network to ensure a fair comparison.
\method significantly improves instance segmentation performance, enhancing overall mAP by +6.8\%.
PartNet benchmark algorithm~\cite{mo2019partnet} and \method both divide points to disjoint instance masks conceptually. However, each instance mask in the former method has no explicit meaning and is optimized by the Hungarian algorithm, while in \method it corresponds to a predefined instance category modeled by position information. As a reward, the optimization objective becomes explicit and easier to realize.

Fig.~\ref{fig:partnet_vis} exhibits qualitative visualization results on PartNet, where ICM-3D demonstrates great capability to segment fine-grained and heterogeneous parts.
We also find some object parts are indistinguishable under \method, as shown in `Failure Case'. The similarities among them are that they own highly close centroids or intertwine with each other. 
It becomes an obstacle for \method to separate parts, as these parts' ground-truth instance category labels may be identical. Therefore we could explain why \method achieves slightly inferior performance on several object categories such as clock and dish, as reported in Tab.~\ref{table-partnet}.
Future work will be focused on handling these failure cases, \textit{e.g.,} using Repulsion Loss~\cite{wang2018repulsion} to repel close objects.

\textbf{OCID.} We follow the settings of UOIS-Net-3D~\cite{xie2021unseen} that trains on synthetic TOD RGB-D dataset~\cite{xie2021unseen} while tests on realistic OCID~\cite{DBLP:conf/icra/SuchiPFV19} to verify model capability on segmenting unseen objects. Because evaluation on OCID requires the network to reason in pixel space, original \method is shifted following DSN module in UOIS-Net-3D by taking as input a 3-channel organized point cloud $\mathcal{P}\in\mathbb{R}^{H\times W\times3}$, of XYZ coordinates, and using its backbone. We accordingly divide pixel space into $N_s\times N_s$ grids for per-pixel classification, and keep the score prediction branch and others the same. 
Experimental results are shown in Tab.~\ref{tab:ocid} where \method achieves very inspiring performance compared to state-of-the-art UOIS-Net-3D. We conjecture that two characteristics of ICM-3D lead to the great generalization to unseen scenes. 1) Using position information to model instance categories can mitigate the reliance on object shapes and semantic classes, making the model more robust to object diversity; 2) Classifying each point to an instance category abandons the clustering post-processing and thereby avoids many dataset-specific hyperparameters.

\begin{figure*}[t]
\vspace{10pt}
\begin{minipage}{0.32\textwidth}
\centering
\includegraphics[width=0.92\textwidth]{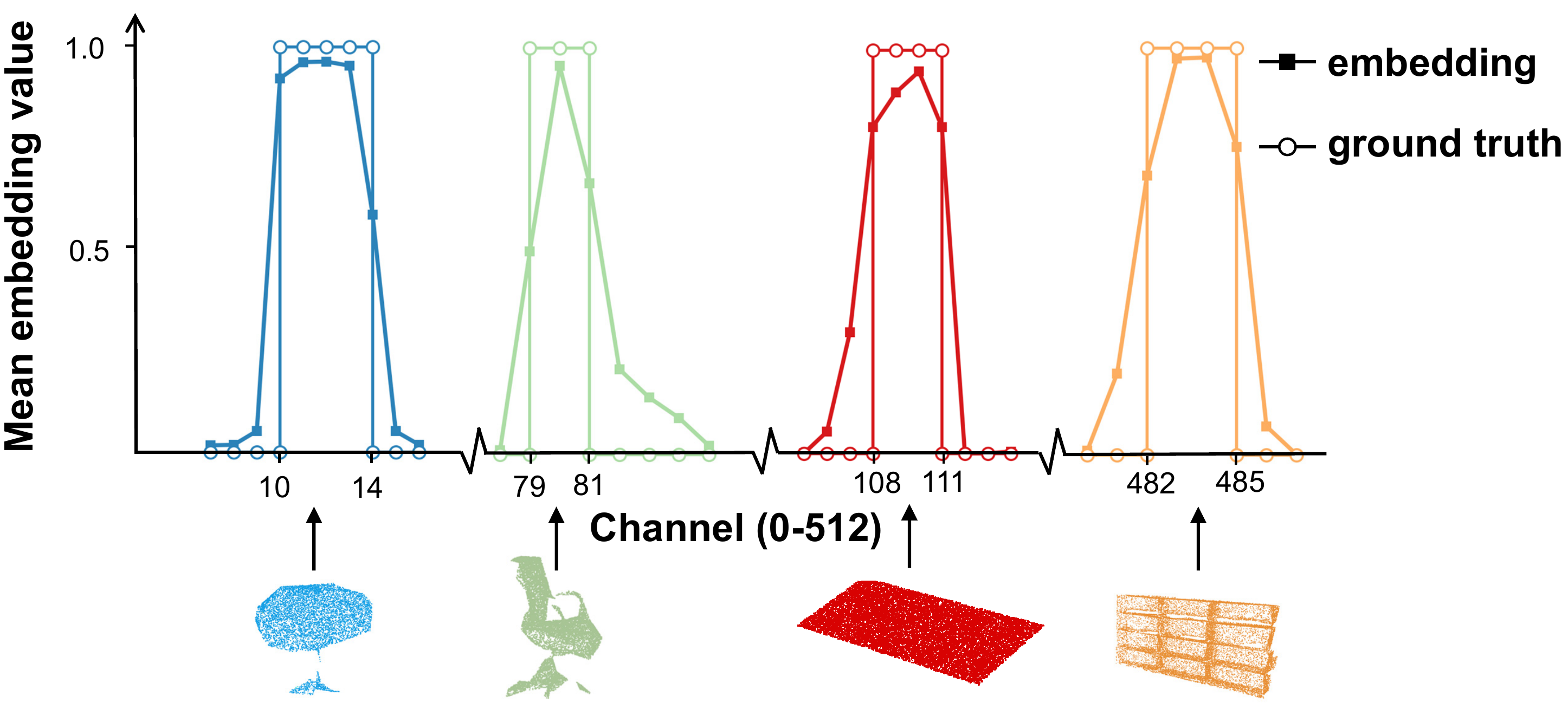}
\caption{Mean embedding values of instances of \method. 
It showcases 4 instances are activated at corresponding channels, thus being distinguished.
}
\label{fig:embedding}
\end{minipage}%
\hspace{5mm}
\begin{minipage}{0.32\textwidth}
\centering
\includegraphics[width=0.92\textwidth]{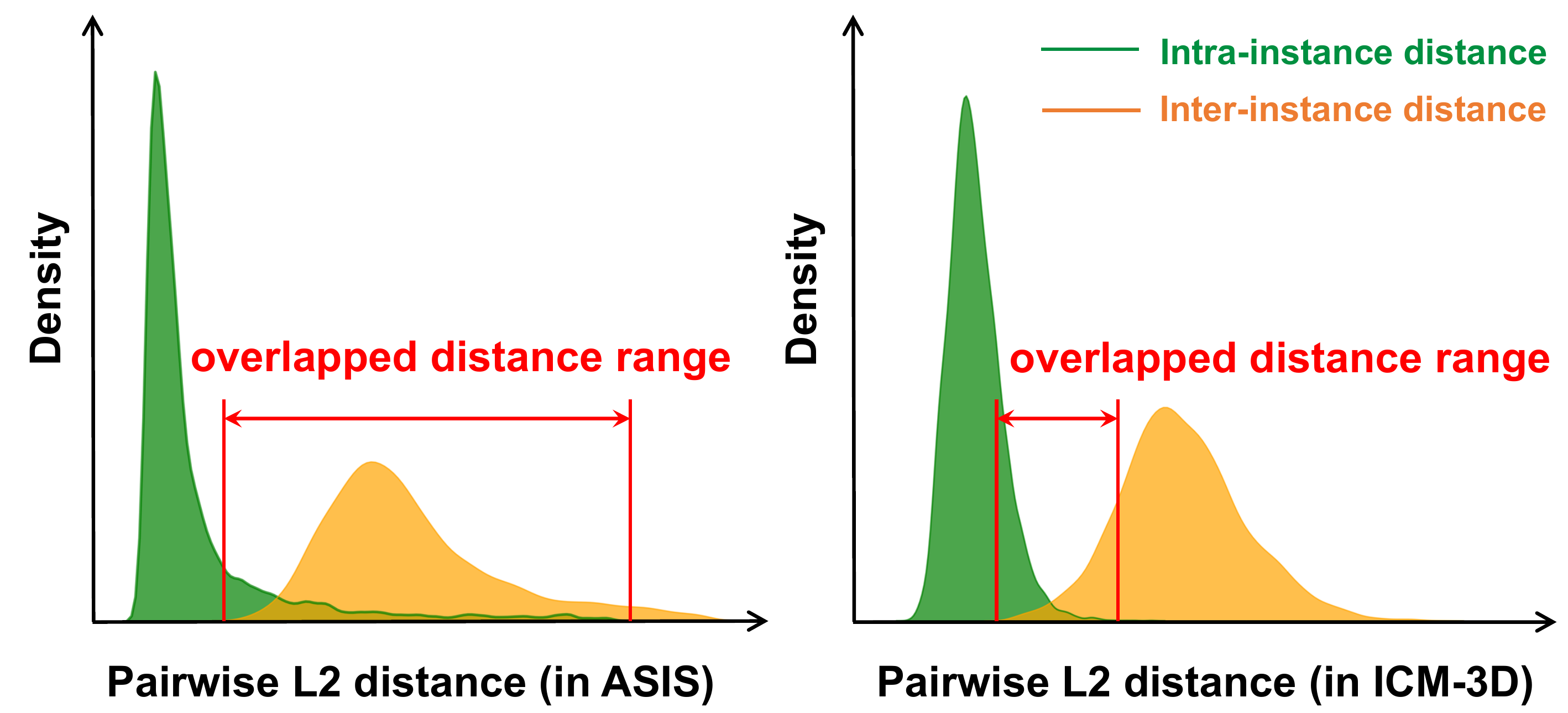}
\caption{Distance distribution comparison. The smaller overlapped distance range indicates the better separation of inter-instance features.}
\label{fig:distance}
\end{minipage}
\hspace{5mm}
\begin{minipage}{0.27\textwidth}
\centering
\includegraphics[width=0.92\linewidth]{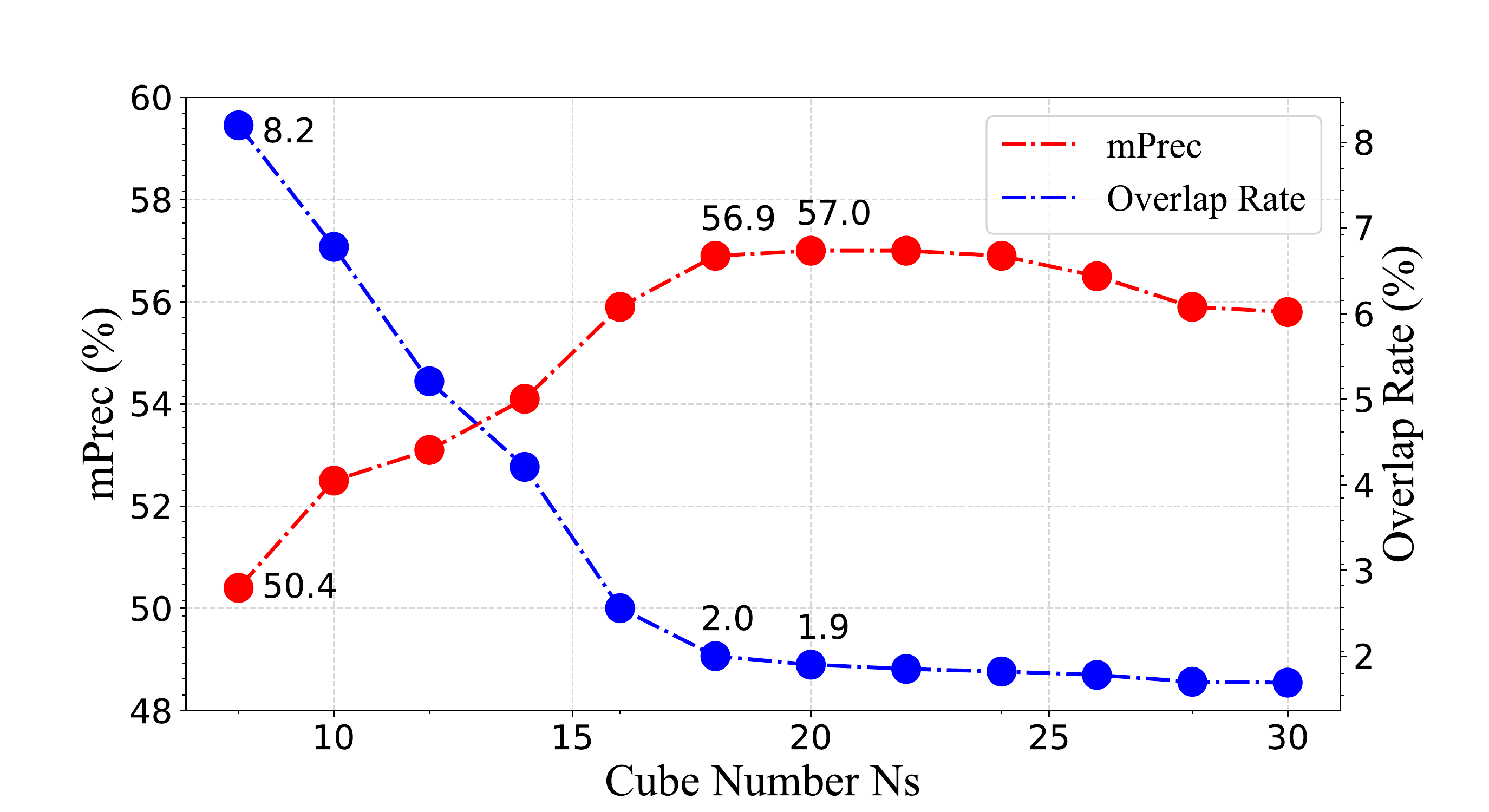}
\caption{The performance and cube overlap rate change with $N_s$. We set $N_s=20$ in other experiments.}
\label{fig:ns}
\end{minipage}
\vspace{-10pt}
\end{figure*}

\subsection{How \method Works}

To investigate how \method works, we visualize the embedding output generated by our method. 
$N_s$ is set as 8 for better visualization. 
For points of the same instance, we calculate their mean embedding values on each channel respectively and plot a value-channel curve, as shown in Fig.~\ref{fig:embedding}. 
The mean embeddings of different instances (denoted as different colors) have the peak value at non-overlapped channels, respectively matching their instance categories modeled by spatial cubes. It enables points of different instances explicitly to be separated in a classification way.

To further compare ICM-3D and the \textit{embed-and-cluster} approach, we investigate the feature distance distribution that could embody feature discriminability.
In Fig.~\ref{fig:distance}, we compute pairwise L2 distances of embeddings learned by \method and ASIS, respectively. 
Although ASIS directly optimizes point distances at training stage, 
the distance ranges of intra-instance points and inter-instance points in ASIS are more highly overlapped. In \method, points of the same instance are squeezed to a smaller range (smaller distances), and thus are distinctive from other points. More precisely, the overlapped probability, \textit{i.e.,} the proportions of overlapped areas, in ICM-3D and ASIS are 2.35\% and 6.89\%, respectively. The statistics imply that by per-point classification, \method well limits the boundaries for points in embedding space, making the instance output of high objectiveness. It remains consistent with the observation in Fig.~\ref{fig:s3dis_vis} that ICM-3D owns superior performance on boundary segmentation.

\subsection{Ablation Study}
\subsubsection{Impact of Instance Category Granularity}
\label{subsubsec: granularity}

The cube number $N_s$ is critical in \method that determines the granularity of instance categories. Theoretically, with a small $N_s$, the central regions of different instances, especially small objects, are more likely to be incorporated in one cube. It will wrongly assign more than one instance category to a cube and thus harm the model accuracy. To verify this, we evaluate the cube overlap rate along with the performance among different $N_s$. Specifically, the cube overlap rate refers to the proportion of cubes that matches multiple instance categories, $N_s$ is varied from 8 to 28 with a stride of 2. We test on S3DIS Area5 set with mPrec as the protocol.

Experimental results shown in Fig.~\ref{fig:ns} confirms that $N_s$ influences the segmentation accuracy by mainly changing the cube overlap rate.
If we choose $N_s$ as 8, the cube overlap rate can reach 8.2\% and the overlapped cases are mainly from small objects, as we define, the longest side of whose bounding box is shorter than 20 cm. Specifically, 15\% small objects cannot be discriminated. When $N_s$ increases to 18, the cube overlap rate drops rapidly to 2.0\% and simultaneously the mPrec gains a considerable increase (from 50.4\% to 56.9\%). Then both remain steady until $N_s$ goes up to 24. In the interval $N_s\in[18, 24]$, we observe that only 4\% small objects are lost, which significantly improves the instance prediction accuracy. Regarding the slight performance drop caused by a larger $N_s$ (greater than 24), we think too much granularity reduces the training sample for each instance category. Consequently, the classification projection to some categories 
is insufficiently learned. This experiment guides us to easily select $N_s$ according to the point cloud distribution, \textit{e.g.,} on object sizes, of a given dataset. In our paper, we choose $N_s$ = 20 due to the best network performance.  
  
\begin{table}[htbp]
 \caption{Relative Performance of the project paradigm compared to the flatten paradigm.}
	\begin{center}
		\renewcommand\arraystretch{1.1}
		\setlength{\tabcolsep}{2.0mm}{
\begin{tabular}{lcccccc}
\toprule
     $N_s$ & 8 & 12 & 16 & 20 & 24 & 28\\
\hline
     mPrec & -1.43 & -1.05 & {-1.03} & {-0.98} & {-0.55} & \textcolor[RGB]{34,139,34}{+0.12}\\
     mRec & {-0.37} & {-0.28} & {-0.21} & {-0.21} & \textcolor[RGB]{34,139,34}{+0.14} & \textcolor[RGB]{34,139,34}{+0.23}\\
     mAP & {-0.41} & {-0.30} & {-0.25} & {-0.19} & \textcolor[RGB]{34,139,34}{+0.10} & \textcolor[RGB]{34,139,34}{+0.18}\\
\bottomrule
\end{tabular}}
 \end{center}
 \label{tab:twopara}
\end{table}

\subsubsection{Comparison of Two Paradigms}

Since two paradigms are proposed to implement \method,
we compare their performance on S3DIS Area5 set and PartNet with various $N_s$. To better visualize the difference, the relative performance is reported in Tab.~\ref{tab:twopara}. Specifically, the flatten paradigm has slight superiority on three metrics with a relatively small $N_s$. We guess that in the project paradigm, decoupling three projection positions from fused high-level point features could be difficult sometimes, giving rise to the marginal performance gap. Notably, this margin narrows and even reverses with a larger $N_s$. For the flatten project, the cubic growth of instance categories leads to a decrease of training samples averaged to per-instance category (as mentioned in Sec.~\ref{subsubsec: granularity}). In contrast, the project paradigm gets rid of this side effect much more. Considering both the competitive performance and substantial memory savings, the project paradigm enables \method to be a highly practical method. 

\subsubsection{Ablation on instance category scoring head} 
We propose an instance category scoring head that evaluates the confidence probability of each cube being valid (see Sec.~\ref{sec:networks}). During the inference, instance categories can be ranked for NMS. An alternative way is to simply use the averaged semantic probability of points belonging to an instance as the quality confidence. By this means, the results in terms of mPrec/mRec on S3DIS Area5 set are dropped by 4.9\%/5.8\%. It indicates that our module is critical for improving the performance by providing precise cube scores for NMS.

\subsection{Test Time}
Replacing traversal-based clustering process with a parallel-computing classifier benefits ICM-3D much for the model efficiency. Statistically, ICM-3D with the project paradigm only consumes 12 ms to process 4k points, 1.4× faster than the ﬂatten paradigm (17 ms). It also enjoys much shorter run-time than both top-down approaches (\textit{e.g.,} 1.6× faster than 3D-BoNet) and competitive embed-and-cluster approaches (\textit{e.g.,} 17× faster than ASIS and 1.4× faster than PointGroup). All models are fairly evaluated under the PointNet++ backbone on a same Tesla-V100 GPU. The low latency enables \method to unleash the potential for segmenting instance-level objects at real time.

\subsection{Difference from SOLO~\cite{wang2019solo} in 2D domain}
\label{sec:solo}

2D instance segmentation method SOLO~\cite{wang2019solo} is related to our \method as it refers to object locations and sizes for pixel classification. Nevertheless, there are significant challenges before making this idea work in 3D domain. (a) Data representation: 2D pixels are uniform and dense, while 3D point clouds are disordered and sparse. Directly adopting 2D data processing manner into 3D, \textit{i.e.,} applying a classifier on regular voxelized 3D grids, will fail to leverage sparsity in the data and suffer from high computation cost. Here in ICM-3D, we process a point-wise classification to preserve geometric details. (b) Context feature extraction: both ICM-3D and SOLO predict instance probabilities based on the object-centric context. In 2D images, it can be easily captured by typical CNNs, as the object center is surrounded by continuous pixels. 
However, 3D object centers are likely to be in empty space, far from any point. 
So ICM-3D dynamically aggregates context features of the object center based on space vicinity (see Sec.~\ref{sec:networks}), making the local pattern adaptive to point densities and object sizes.
As a result, the probability scores become more reliable. Simply extending SOLO to 3D domain, \textit{i.e.,} 3D grid classification \& fixed context aggregation, results in much worse performance (-16.9\% mPrec on S3DIS) compared to ICM-3D.

\section{CONCLUSIONS}

In this paper, we reformulate the 3D instance segmentation into a per-point classification problem by introducing the concept of `instance category'. 
Compared with previous \textit{embed-and-cluster} paradigm, our method directly separate instances by predicting their corresponding categories, thus eliminating the additional clustering post-processing. Despite its simplicity, our instantiated category modeling paradigm shows its priority on various datasets and backbone models. We also conduct an extensive analysis to show how and why it works. 
We hope the novel design provides valuable insights to future works on 3D point cloud analysis, \textit{e.g.}, instance segmentation, and beyond.


%



\ifCLASSOPTIONcaptionsoff
  \newpage
\fi



\bibliographystyle{IEEEtran}
\bibliography{IEEEabrv,reference}
%

%





\end{document}